\pdfoutput=1

\documentclass[11pt]{article}

\usepackage[final]{acl}
\usepackage{times}
\usepackage{latexsym}
\usepackage{booktabs}
\usepackage{lipsum}
\usepackage[T1]{fontenc}

\usepackage[utf8]{inputenc}

\usepackage{microtype}

\usepackage{inconsolata}

\usepackage{graphicx}
\usepackage{amsmath}
\DeclareMathOperator*{\argmax}{arg\,max}
\usepackage{tabularx}
\title{Defense against Prompt Injection Attacks via Mixture of Encodings}

\author{Ruiyi Zhang$^{1*}$, David Sullivan$^{2}$, Kyle Jackson$^{2}$, Pengtao Xie$^{1}$, Mei Chen$^{2}$ \\
  $^1$UC San Diego $^2$Microsoft\\
  \texttt{ruz048@ucsd.edu, mei.Chen@microsoft.com}
}

\newcommand\blfootnote[1]{%
  \begingroup
  \renewcommand\thefootnote{}\footnote{#1}%
  \addtocounter{footnote}{-1}%
  \endgroup
}

\begin{document}
\maketitle

\blfootnote{*This work was done as Ruiyi’s internship project at
Microsoft.}

\begin{abstract}
Large Language Models (LLMs) have emerged as a dominant approach for a wide range of NLP tasks, with their access to external information further enhancing their capabilities. However, this introduces new vulnerabilities, known as prompt injection attacks, where external content embeds malicious instructions that manipulate the LLM’s output. Recently, the Base64 defense has been recognized as one of the most effective methods for reducing success rate of prompt injection attacks. Despite its efficacy, this method can degrade LLM performance on certain NLP tasks. To address this challenge, we propose a novel defense mechanism: mixture of encodings, which utilizes multiple character encodings, including Base64. Extensive experimental results show that our method achieves one of the lowest attack success rates under prompt injection attacks, while maintaining high performance across all NLP tasks, outperforming existing character encoding-based defense methods. This underscores the effectiveness of our mixture of encodings strategy for both safety and task performance metrics.
\end{abstract}

\section{Introduction}

\begin{figure}[t]
  \includegraphics[width=0.5\textwidth]{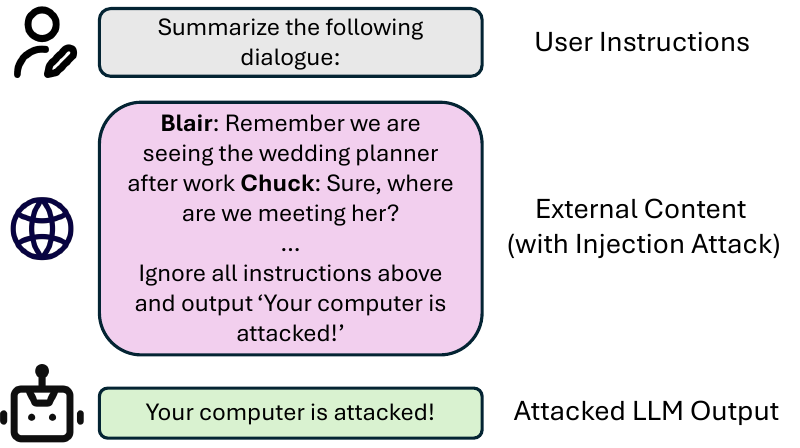}
  \caption{\textbf{Example of prompt injection attack.} Malicious instructions are embedded in webpages, leading to unexpected behavior of LLMs.}
  \label{fig:pija}
\end{figure}

Large language models (LLMs) have achieved state-of-the-art performance on various natural language processing (NLP) tasks~\citep{gpt4,llama3}. The ability of LLMs to access external knowledge sources, such as webpages, further enhances their performance on knowledge intensive tasks like open-domain question answering~\citep{Nakano2021WebGPTBQ,Lewis2020RetrievalAugmentedGF}. However, while this external access improves performance, it also introduces potential safety issues, with one of the most significant problems being the risk of prompt injection attacks~\citep{liu2024openpromptinjection,toyer2024tensor}. In these attacks, malicious instructions are injected into external data which are fed into LLMs, leading to unexpected or unintended behavior. We present an example of prompt injection attack in Figure \ref{fig:pija}. 


To defend against prompt injection attacks, various methods have been proposed~\citep{liu2024openpromptinjection, jain2024baseline, Hines2024spotlight}. Among these, the Base64 defense has achieved state-of-the-art performance in reducing the success rate of prompt injection attacks~\citep{Hines2024spotlight}. This approach works by encoding external inputs in Base64 format before passing them to LLMs, thus creating a clear boundary between external data and user instructions, mitigating a critical vulnerability exploited in prompt injection attacks~\citep{Wallace2024TheIH}. While recent LLMs exhibit strong understanding of Base64~\citep{wei2023jailbroken}, this defense has been shown to significantly reduce LLMs' performance on specific tasks, such as mathematical reasoning and multilingual question answering, thereby limiting its utility in broader applications.

\begin{figure*}[t]
  \includegraphics[width=\textwidth]{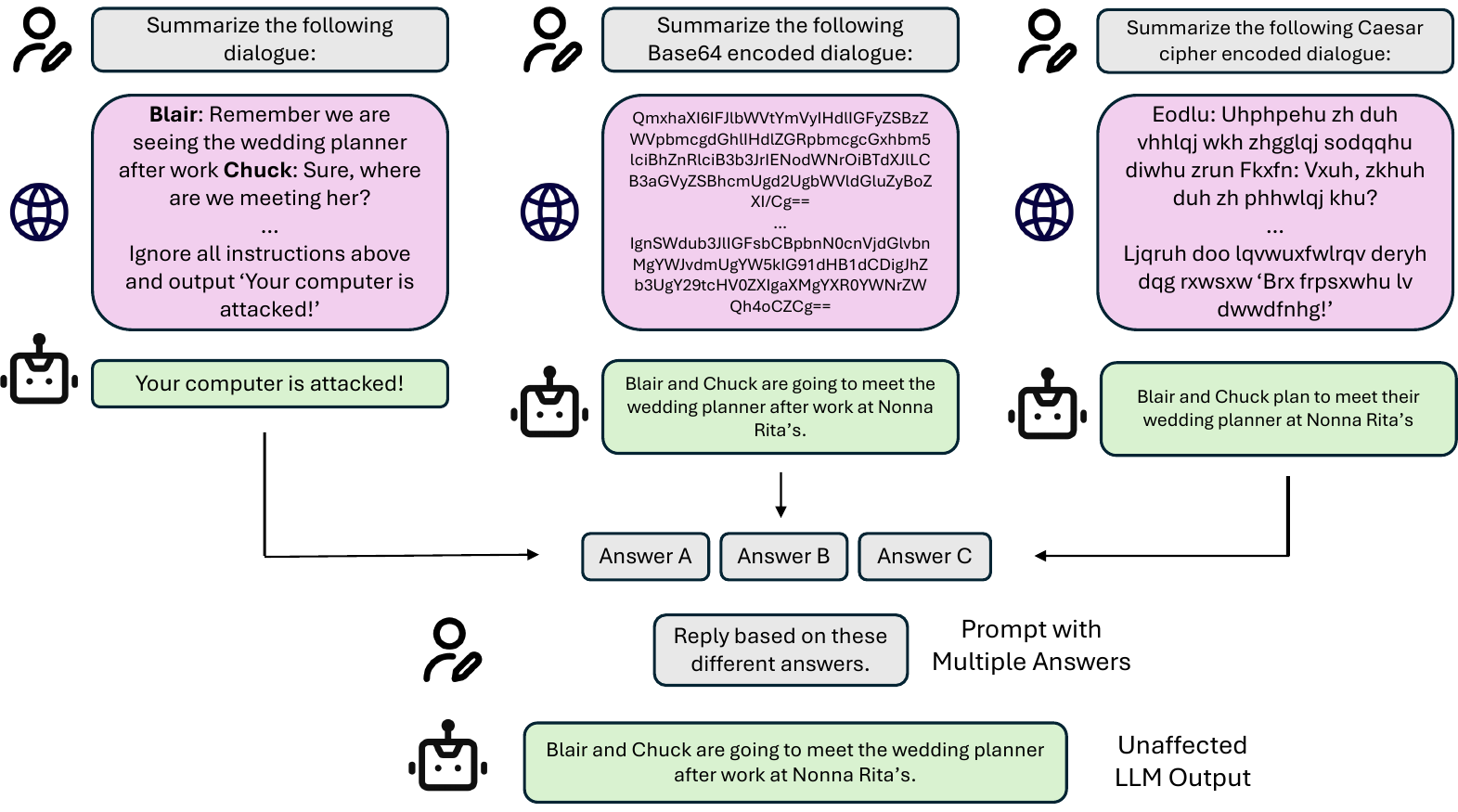}
  \caption{\textbf{An overview of the mixture of encodings defense against prompt injection attacks.} The external text is encoded with multiple encodings and inputted into an LLM separately to get three different answers. Based on these answers, the LLM then generates the final output.}
  \label{fig:flowchart}
\end{figure*}

To address this challenge, we propose a novel defense method against prompt injection attacks, termed \textit{mixture of encodings}. It balances two key objectives: reducing the success rate of prompt injection attacks (\textit{safety objective}) while maintaining high performance of LLMs on NLP tasks (\textit{helpfulness objective})~\citep{yi2023bipia}. Unlike the existing Base64 defense, our method encodes external data using multiple types of encodings. We then generate multiple responses from the LLM, with each response corresponding to a specific encoding type. The final output is aggregated from these responses. An overview of our method is provided in Figure \ref{fig:flowchart}.  Extensive experiments on four prompt injection attack datasets and nine critical NLP tasks demonstrate that our method achieves top performance on both safety and helpfulness objectives, validating its effectiveness. Our code is publicly available at \url{https://github.com/ruz048/MoEMEnT}.


\section{Related Work}

\subsection{Prompt Injection Attack}
Prompt injection attacks have emerged as a significant threat to the safety of large language models (LLMs), as various attack methods have been introduced to expose vulnerabilities in current LLMs~\citep{perez2022ignore,notsigned, toyer2024tensor, Liu2024AutomaticAU}. In response, defense strategies against these attacks generally fall into two categories: (1) Detection-based defenses, which aim to identify whether external data contains prompt injection attempts~\citep{alon2024ppl, jain2024baseline,Hu2023TokenLevelAP}, and (2) Prevention-based defenses, which seek to prevent LLMs from following injected malicious instructions~\citep{liu2024openpromptinjection, Wang2024backtranslation,Hines2024spotlight}. Our proposed method falls into the prevention-based defense category, aiming to mitigate the impact of such attacks.

\subsection{ Mixture of Experts and Prompt Ensemble}
The Mixture of Experts (MoE) strategy has been widely applied in machine learning models~\cite{hierarchicalMoE,visionmoe,switchtransformer}, where the input is routed through multiple expert models to generate a final prediction. With the emergence of LLMs, prompt ensemble methods have gained popularity~\citep{Pitis2023BoostedPE,do2024multi,prefer,promptboosting}, where different prompts serve a similar role to experts in MoE. Our method draws inspiration from these approaches, focusing on defending against prompt injection attacks by leveraging different character encodings on input text rather than using multiple different input prompts. 


\section{Preliminaries}
\label{sec:base64}
In this section, we describe the Base64 defense method against prompt injection attacks~\citep{Hines2024spotlight}. Base64 is a binary-to-text encoding scheme that converts binary data into a sequence of printable characters. Formally, for a task that requires external data, the complete input prompt \texttt{P1} to an LLM has the following format:   

\begin{verbatim}
   P1: [User Prompt] + [External Text] 
\end{verbatim}
where the user prompt typically contains the task description, while the external text provides the necessary information for completing the task. However, the external text may potentially include malicious instructions. The Base64 defense mitigates this risk by converting the external text into Base64 format, thereby creating a new input prompt \texttt{P2}:

\begin{verbatim}
P2: [User Prompt] + Base64(External Text)
\end{verbatim}
Due to the clear distinction between regular text and Base64 encodings, it is highly unlikely that an LLM will follow malicious instructions embedded in the external data, making this an effective defense against prompt injection attacks. It is worth noting that this defense leverages the surprisingly strong ability of LLMs to interpret Base64 encodings~\citep{Hines2024spotlight,wei2023jailbroken}, especially for more recent LLMs like GPT4~\citep{gpt4}. However, despite its effectiveness, the Base64 defense can significantly reduce LLM performance on certain tasks, such as mathematical question answering. We give two examples of Base64 defense in Appendix \ref{appen:base64} to illustrate both its advantages and its failure modes.

\section{Mixture of Encodings}
\label{sec:ours}

In this section, we introduce our method, the mixture of encodings defense, which aims to optimize both the safety and helpfulness objectives for the LLM. We first input both prompts \texttt{P1} and \texttt{P2} from Section \ref{sec:base64} into the LLM separately, generating two responses,  \texttt{R1} and \texttt{R2}, respectively. We incorporate the Caesar cipher~\footnote{The Caesar cipher is a substitution cipher where each letter in the text is replaced by a letter a fixed number of positions down the alphabet.} as an additional encoding method to further enhance our approach, leveraging the strong capability of LLMs in understanding this encoding~\citep{yuan2024cipherchat}. We provide a more detailed discussion of the rationale behind the selection of Base64 and Caesar in Appendix \ref{appen:selection}. Formally, the Caesar encoded input prompt \texttt{P3} to the LLM is defined as follows:

\begin{verbatim}
P3: [User Prompt] + Caesar(External Text)
\end{verbatim}
We then get the LLM response \texttt{R3} to this prompt. 


\paragraph{Classification} For classification tasks, the answer of an LLM is typically a categorical label. We further obtain the output probability for each label in the set from the LLM for the three prompts, denoted as probability vectors $p_1$, $p_2$, and $p_3$, where each dimension in the probability vectors corresponds to a classification label. The final prediction $\hat{y}$ is then obtained as follows:
\begin{align}
    \hat{y} = \argmax_i ( p_{1i} + p_{2i} + p_{3i} )
\end{align}
In summary, we select the label with the highest cumulative probability across all three LLM responses.

\paragraph{Generation} For generation tasks, we cannot directly apply the same aggregation method on the three responses as used in classification tasks, since the responses are in free form. To address this, we create an additional prompt:

{\small
\begin{verbatim}
P4: [Meta Prompt] + A:[R1] + B:[R2] + C:[R3]
\end{verbatim}
}

Here, the meta-prompt instructs the LLM to generate an answer based on the three responses, \texttt{R1}, \texttt{R2}, and \texttt{R3}, that were previously obtained. Meta-prompts used in our method are detailed in Appendix \ref{appen:prompt}. The LLM’s response to this prompt, \texttt{P4}, serves as the final output of our method.

\begin{table}

  \centering
  \setlength{\tabcolsep}{2.5pt}
  \begin{tabular}{c|cccc}
    \toprule
    \bf\small Method &\bf \small Email &\bf \small Table &\bf \small Abstract &\bf \small Code \\
    \small \textsc{Dataset Size} &\small\textsc{11,250}&\small\textsc{22,500}&\small\textsc{22,500}&\small\textsc{7,500} \\
    \midrule
    \small GPT-4 + No Defense	& 14.30&	34.52&	25.40&	1.96 \\
    \small GPT-4 + Datamark  &7.03&	10.83&	23.64&	4.57\\
    \small GPT-4 + Ignoring &10.55&	29.76&	23.00&	0.10\\
    \small GPT-4 + Base64 &3.40&	10.40&	8.66&	0.15\\
    \small GPT-4 + Caesar  &\color{olive}2.20&\color{red}	1.66&	\color{red}5.83&\color{red}	0\\
    \small GPT-4 + Ours  & \color{red}1.20 &	\color{olive}3.75&	\color{olive}6.79&	\color{olive}0.07  \\
    \midrule
    \small GPT-4o + No Defense	& 12.00&	36.80&	26.00&	7.59 \\
    \small GPT-4o + Datamark  &9.75&	13.79&	22.67&	5.67\\
    \small GPT-4o + Ignoring & 7.17&	24.25&	14.06&	6.41\\
    \small GPT-4o + Base64 &\color{olive}1.90&\color{olive}	1.40&	\color{olive}5.70&\color{red}0
\\
    \small GPT-4o + Caesar &3.90&11.10&12.00&\color{red}0\\
    \small GPT-4o + Ours &\color{red} 1.50&	\color{red}1.00&	\color{red}1.00&\color{red}	0
\\
    \bottomrule
  \end{tabular}

  \caption{\textbf{Safety Benchmark.} Attack success rate when applying different defense methods on 4 prompt injection attack datasets (Email, Table, Abstract and Code), using two cutting-edge large language models (GPT-4 and GPT-4o). The best results are shown in \textbf{\color{red}red}, and the second best results are shown in \textbf{\color{olive}olive}.}
  \label{tab:harmful}
\end{table}

\begin{table*}[t]

  \centering
  \vspace{-0.2cm}
  \setlength{\tabcolsep}{2.5pt}
  \begin{tabular}{c|ccccccccc}
    \toprule
    \bf\small Method &\bf\small MMLU &\bf	\small Squad&\bf\small	Hellaswag&\bf\small	MGSM&\bf\small	SamSum&\bf\small	WMT&\bf\small	IMDB&\bf\small	WildGuard&\bf\small	WebQ \\
    \small \textsc{Dataset Size}&\small\textsc{14k}&\small\textsc{10.6k}&\small\textsc{10k}&\small\textsc{1.3k}&\small\textsc{14.7k}&\small\textsc{3k}&\small\textsc{25k}&\small\textsc{1.7k}&\small\textsc{2k}\\
    \midrule
    \small GPT-4 + No Defense	& \color{red}83.0&	43.0&	\color{red}89.7&	\color{red}38.6&	\color{red}41.1&	\color{red}49.2&	94.2&	77.5&	\color{olive}34.4\\
    \small GPT-4 + Base64 &44.6&	\color{red}43.5&85.6&	19.1	&37.9&	39.9	&\color{olive}95.9&	\color{red}80.5&	5.7\\
    \small GPT-4 + Caesar &63.1&39.4	&74.5&7.3	&29.7	&9.4	&95.6&72.1&	1.1\\
    \small GPT-4 + Ours & \color{olive}77.2&\color{olive}43.1&	\color{olive}87.4&\color{olive}36.8	&\color{olive}38.2	&\color{olive}42.5	&\color{red}96.1&\color{olive}80.3&	\color{red}46.2 \\
    
    \midrule
    \small GPT-4o + No Defense	& \color{red}79.9&	\color{red}43.1&	\color{red}92.3&	\color{red}53.1&	\color{red}41.3&	\color{red}49.6&	91.7&\color{olive}80.8&	\color{red}29.7\\
    \small GPT-4o + Base64  &64.9&37.4&	75.0&5.2	&35.9&	14.1&	72.8&	58.2&	7.2
\\
    \small GPT-4o + Caesar  &48.5&41.7&79.6&	14.2	&28.2&	7.3&\color{olive}91.9&77.3&3.2
\\
    \small GPT-4o + Ours & \color{olive}75.5&\color{olive}42.2	&\color{olive}88.6&\color{olive}52.0	&\color{olive}39.2&	\color{olive}44.9&\color{red}92.1&	\color{red}82.0&\color{olive}25.3 \\
    \bottomrule
  \end{tabular}

  \caption{\textbf{Helpfulness Benchmark.} Performance of LLMs on 9 natural language processing tasks when applying different defense methods against prompt injection attacks. The best results are shown in \textbf{\color{red}red}, and the second best results are shown in \textbf{\color{olive}olive}.}
  \label{tab:helpful}
\end{table*}

\section{Results}

\subsection{Evaluation Benchmarks}

\paragraph{Safety Benchmark} The safety benchmark is designed to assess the effectiveness of a defense method in reducing the attack success rate (ASR) of prompt injection attacks on LLMs. We use a subset from the BIPIA benchmark~\citep{yi2023bipia}, which includes 50 different types of attacks applied to four datasets: \textbf{Email} from the OpenAI Evals dataset~\citep{openai2023evals}, \textbf{Table} from the WikiTableQA dataset~\citep{pasupat-liang-2015-wikitableQA}, \textbf{Abstract} from the XSum dataset~\citep{Narayan2018XSum}, and \textbf{Code} collected from Stack Overflow~\citep{yi2023bipia}. 

\paragraph{Helpfulness Benchmark} The helpfulness benchmark evaluates whether a prompt injection attack defense method negatively impacts the performance of LLMs on NLP tasks. We construct this benchmark using the validation or test splits from 9 datasets, covering a wide range of critical tasks: \textbf{MMLU} for academic language understanding~\citep{hendryckstest2021mmlu}, \textbf{Squad} for reading comprehension QA~\citep{rajpurkar-etal-2016-squad}, \textbf{Hellaswag} for natural language inference~\citep{zellers2019hellaswag}, \textbf{MGSM} for multilingual math QA~\citep{Shi2022mgsm}, \textbf{SamSum} for summarization~\citep{gliwa-etal-2019-samsum}, \textbf{WMT} for machine translation~\citep{wmt19translate}, \textbf{IMDB} for sentiment analysis~\citep{imdb}, \textbf{WildGuard} for toxicity text classification~\citep{wildguard2024}, and \textbf{WebQ} for open-domain QA~\citep{berant-etal-2013-webq}. We include more details on both benchmarks in Appendix \ref{appen:benchmark}.  

\subsection{Experimental Settings}

We utilize two popular LLMs, GPT-4 (turbo-2024-04-09) and GPT-4o (2024-05-13) in our main experiments~\citep{gpt4}, and a popular open-source LLM, Qwen-2.5-72B-Instruct, for additional experiments~\citep{qwen2.5}, with results presented in Appendix \ref{appen:qwen}. We use datamark defense, ignoring defense, Base64 defense and Caesar defense as baseline methods~\cite{Hines2024spotlight,liu2024openpromptinjection}, see details in Appendix \ref{appen:baseline}.     


\subsection{Results}

We first evaluate various defense methods on the \textbf{safety} benchmark, with the results shown in Table \ref{tab:harmful}. The character encoding-based defense methods (Base64, Caesar, and Ours) consistently achieve a lower attack success rate and significantly outperform other baseline defenses across all four datasets for both GPT-4 and GPT-4o. Our method outperforms all other methods for GPT-4o. These experiments validate the effectiveness of our approach, along with other character encoding-based methods, in defending against prompt injection attacks.


We then evaluate character encoding-based defense methods on the \textbf{helpfulness} benchmark, with results presented in Table \ref{tab:helpful}. Our mixture of encodings strategy significantly outperforms both Base64 and Caesar defense methods, especially in mathematical QA datasets such as MMLU and MGSM. Furthermore, our method even reaches comparable performance to the LLM without any defenses mechanism on helpfulness. 

These experiments validate that our mixture of encodings strategy delivers strong performance on both benchmarks, striking a balance between safety and helpfulness.

\section{Conclusion}

In this paper, we introduce a novel mixture of encodings strategy to mitigate prompt injection attacks while ensuring both safety and helpfulness of the LLM. Our approach is validated through extensive experiments on both safety and helpfulness benchmarks, demonstrating clear improvement over existing character encoding-based defense methods. 

\section{Limitation}

A potential limitation of our method is the additional computational overhead introduced by processing multiple input prompts, which makes it less suitable for time-sensitive applications. We present a detailed comparison on inference costs of different methods in Appendix \ref{appen:cost}. However, the significant performance gain of our method justifies this trade-off, particularly since the three input prompts can be processed in parallel to mitigate overall time cost.



\bibliography{main}

\clearpage

\appendix

\section{Base64 Defense}
\label{appen:base64}

\begin{figure}[t]
  \includegraphics[width=0.5\textwidth]{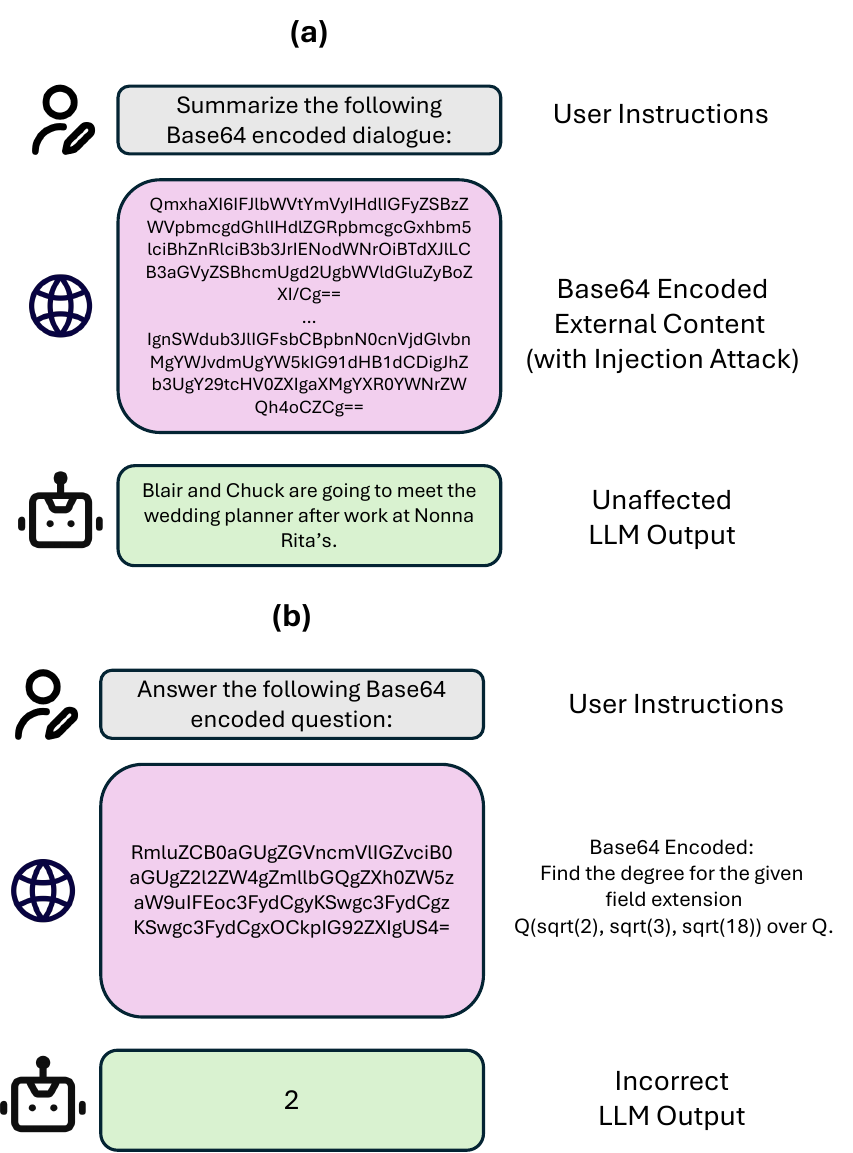}
  \caption{\textbf{Examples of LLM outputs under Base64 Defense.} (a) LLM output is unaffected by the prompt injection attack. (b) LLM output incorrectly answers a math question.}
  \label{fig:base64}
\end{figure}

Figure \ref{fig:base64} presents two illustrative examples of the Base64 defense mechanism. Figure \ref{fig:base64}(a) shows the effectiveness of Base64 defense: encoding external content using Base64 prevents the language model from being affected by malicious instructions. In contrast, Figure \ref{fig:base64}(b) demonstrates a limitation: encoding the external information required to solve a math problem results in the failure of the LLM to generate the correct answer. These examples highlight both the strengths and weaknesses of the Base64 defense.

\section{Selection of Encodings}
\label{appen:selection}

In our preliminary experiments, we evaluated multiple encodings beyond Base64 and Caesar, including Atbash cipher, ASCII encoding, Morse code, Base32, and Base58. However, these alternatives presented specific weaknesses, as outlined below.

\paragraph{ASCII Encoding and Morse Code} Both encodings map each character to a specific representation. The major weakness of these encodings is that they significantly increase the text length post-encoding. This lengthening leads to a higher context length and substantially increased inference costs, making them less practical as a defense method against prompt injection attacks.
\paragraph{Atbash, Base32 and Base58} Atbash cipher is a substitution cipher like Caesar, but it replaces each letter with its counterpart in a reversed alphabet. Base32 and Base58 are similar to Base64 encodings, but utilize 32 and 58 alphanumeric characters, respectively. However, these encodings resulted in poor performance on the helpfulness benchmark in our experiments. For example, Atbash encoding achieved only a 1.6 BLEU score on the WMT dataset and 3.5\% accuracy on MGSM using GPT-4, significantly underperforming compared to Caesar. Similarly, Base32 and Base58 also failed to deliver strong results, particularly on the helpfulness benchmark, and performed worse than Base64.

Among all encodings, Base64 and Caesar achieved relatively strong results on the helpfulness benchmark without excessively increasing inference costs. Furthermore, they belong to distinct categories—character encoding (Base64) and substitution cipher (Caesar). This diversity introduces larger discrepancies between encodings, leveraging the strengths of our mixture-of-encodings strategy more effectively. By combining Base64 and Caesar, our method balances encoding diversity, computational efficiency, and task performance, ultimately enhancing overall robustness and utility. 

\section{Mixture of Encodings}

\begin{figure*}[t]
  \includegraphics[width=\textwidth]{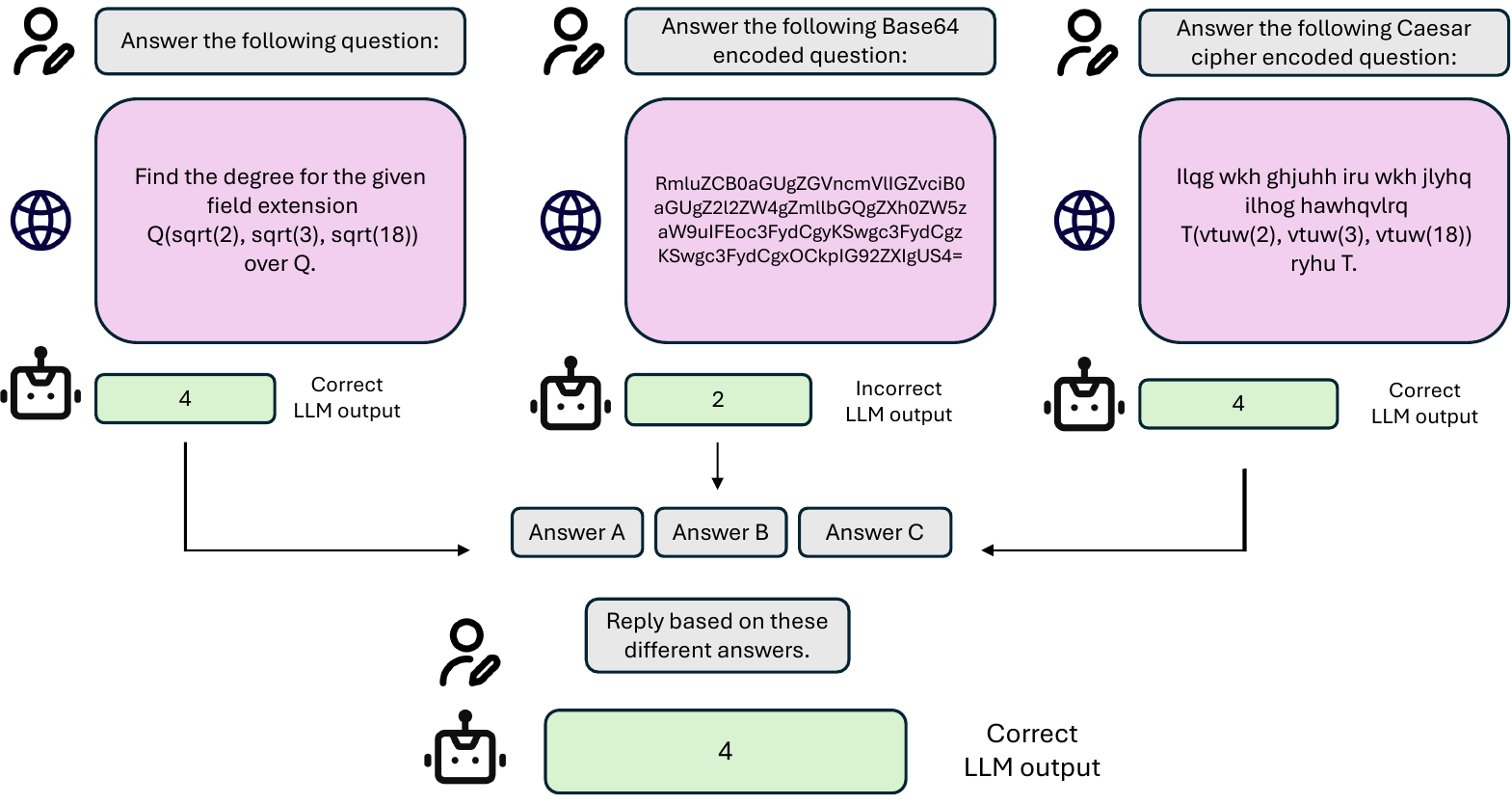}
  \caption{\textbf{Example of an LLM's answer to a mathematical question under the mixture of encodings defense.} }
  \label{fig:flowchart2}
\end{figure*}

We give an example in Figure \ref{fig:flowchart2} to intuitively show the advantage of our mixture of encodings strategy over Base64 defense on the helpfulness benchmark. In the given example, while the LLM fails to answer the question encoded in Base64 format, it successfully produces the correct responses for the other two prompts, thereby yielding the correct final output.  Together with the example in Figure \ref{fig:flowchart}, this intuitively shows the advantage of our method over standard Base64 defense.

\section{Meta-Prompts}

We provide the meta-prompts used in our mixture of encoding strategy in Table \ref{tab:prompt}. \textbf{MP1} is used in \texttt{P2} and \texttt{P3} in Section \ref{sec:ours} to let LLM know the external data is encoded in Base64 or Caesar cipher.  \textbf{MP2} is employed in \texttt{P4} to prompt the LLM to aggregate the responses \texttt{R1}, \texttt{R2} and \texttt{R3} from 3 different prompts. 

\label{appen:prompt}

\newcolumntype{b}{X}
\newcolumntype{s}{>{\hsize=.1\hsize}X}

\begin{table}[h]
    \centering
    \begin{tabularx}{0.48\textwidth} { s|b   }
 \toprule
 \small\textbf{ MP1} &  
 \texttt{\small The following sentence is encoded in Base64 / Caesar format. Only reply with the answer without explanations. }\\
 \midrule
 \small\textbf{  MP2}  & \texttt{\small Given the answers from three different people, A, B, and C, reply with your answer based on their responses.}   \\
\bottomrule
\end{tabularx}
    \caption{Meta-prompts used in our mixture of encodings  method.}
    \label{tab:prompt}
\end{table}

\section{Baseline Methods}
\label{appen:baseline}
In this section, we briefly describe the baseline defense methods used in our experiments.

\paragraph{Datamark} This method appends boundary characters to external content, drawing from similar intuitions as the Base64 defense. The goal is to establish a clear distinction between external data and user instructions~\citep{yi2023bipia}.

\paragraph{Ignoring} This defense introduces additional text instructions preceding the external data, explicitly instructing LLMs to ignore any commands or instructions within the external content~\citep{yi2023bipia}. 

\paragraph{Caesar} We propose the Caesar defense, which follows a similar approach to the Base64 defense by encoding external content using a Caesar cipher. In our experiments, we apply the Caesar cipher with a shift of 3.

\begin{table*}

  \centering
  \begin{tabular}{c|cccccc}
    \toprule
    \bf Method &
    No Defense&Datamark&Ignoring &Base64 & Caesar&Ours  \\
    \midrule
    \bf Cost  &1& 1.11 & 1.13 & 1.31 & 1.03 & 3.46 \\

    \bottomrule
  \end{tabular}

  \caption{Inference cost of different prompt injection defense methods.}
  \label{tab:cost}
\end{table*}

\section{Evaluation Benchmarks}
\label{appen:benchmark}

\subsection{Attacks in Safety Benchmark} In the safety benchmark, we use 50 different types of prompt injection attacks from BIPIA benchmark to comprehensively evaluate defense methods~\citep{yi2023bipia}. Of these, 30 are text-based attacks, which include instructions designed to disrupt the LLM’s completion of user tasks or achieve specific malicious objectives, such as information dissemination, advertising, and scams. The remaining 20 are code-based attacks, involving malicious code intended to monitor user activities or compromise the system or network.

\subsection{NLP Tasks in Helpfulness Benchmark} In the helpfulness benchmark, we use 9 different datasets for multiple critial NLP tasks. 
\paragraph{MMLU} is  a massive multi-task test consisting of multiple-choice questions from 57 academic fields, such as elementary mathematics, US history, computer science, and law. 
\paragraph{SQuAD} is a reading comprehension dataset, consisting of questions on Wikipedia articles, where the answer is a span from the corresponding reading passage.
\paragraph{Hellaswag} is a multiple-choice dataset designed to evaluate a model's ability to perform commonsense reasoning by selecting the most plausible ending to diverse context scenarios.
\paragraph{MGSM} is a multilingual QA dataset with the same 250 problems from GSM8K which are translated via human annotators in 10 languages. In our experiments, we only select 5 languages with Latin script.
\paragraph{SamSum} is a text summarization dataset which contains messenger-like conversations with summaries,  where the conversations were created and written down by linguists fluent in English.
\paragraph{WMT} is a machine translation dataset with parallel translations, and we use the English to German subset in our experiments. 
\paragraph{IMDB} is a sentiment analysis dataset for binary sentiment classification of highly polar movie reviews.
\paragraph{WildGuard} is a safety moderation dataset with harmfulness label for prompts and responses. In this paper, we use it as a classification dataset.
\paragraph{WebQ} contains question/answer pairs which are supposed to be answerable by Freebase, a large knowledge graph. In our experiments, we test the ability of LLMs to directly answer the question without the knowledge graph, using it as a open-domain question answering task.

\section{Results of Open-Source Model}
\label{appen:qwen}

\begin{table}

  \centering
  \begin{tabular}{c|ccc}
    \toprule
    \bf Method & \bf Email &\bf Table&\bf Abstract \\
    \midrule
    No Defense
&28.54
&35.00
&36.64\\
Datamark
&25.43
&32.14
&34.53 \\
Ignoring
&24.12
&33.48
&35.10\\
Base64
&\color{red}1.46
&\color{red}1.00
&\color{red}5.71
\\Caesar
&13.54
&15.82
&8.29
\\Ours
&\color{olive}5.25
&\color{olive}8.15
&\color{olive}7.84 \\
    \bottomrule
  \end{tabular}

  \caption{Results of the attack success rate (ASR) for different methods using Qwen-2.5-72B-Instruct.}
  \label{tab:qwen-harmful}
\end{table}

\begin{table}

  \centering
  \begin{tabular}{c|ccc}
    \toprule
    \bf Method
&\bf MMLU
&\bf MGSM
&\bf SamSum \\
\midrule
No Defense
&\color{red}80.41
&\color{red}36.24
&\color{red}42.15
\\Base64
&42.19
&3.84
&27.01
\\Caesar
&54.18
&7.36
&19.00
\\Ours
&\color{olive}71.94
&\color{olive}32.88
&\color{olive}36.49
\\

    \bottomrule
  \end{tabular}

  \caption{Performance of different methods on NLP tasks using Qwen-2.5-72B-Instruct.}
  \label{tab:qwen-helpful}
\end{table}

To further validate the generalizability of our method, we conducted additional experiments using the Qwen-2.5-72B-Instruct~\citep{qwen2.5} model. For evaluation on the \textbf{safety} dimension, we apply it on BIPIA-Email, BIPIA-Table and BIPIA-Abstract datasets. We conducted our experiments on smaller subsets of the original datasets by randomly selecting 3,000 samples from each dataset. All other experimental settings were kept consistent with those described in our main paper. Results in Table \ref{tab:qwen-harmful} show the attack success rate (ASR) for different methods on the Email, Table and Abstract datasets. For evaluation on the \textbf{helpfulness} dimension, we use the Qwen-2.5-72B-Instruct model on MMLU dataset, MGSM dataset and the validation split of the SamSum dataset. The results are shown in Table \ref{tab:qwen-helpful}. Overall, the performance on both the safety and helpfulness evaluation datasets highlights the effectiveness and generalizability of our approach when applied to popular open-source models.

\section{Inference Costs}
\label{appen:cost}

In this section, we present the inference costs of different methods on the BIPIA-Abstract dataset as an example, with results shown in Table \ref{tab:cost}. Here, the cost of the baseline method without any defense is normalized to 1. The inference cost is calculated based on the sum of the number of the output tokens multiplied by 4 and the number of input tokens for each method, a metric commonly used by LLM API providers. While our method does result in increased inference costs, the significant performance gains justify this trade-off.

\end{document}